\documentclass[10pt,twocolumn,letterpaper]{article}

\usepackage{iccv}
\usepackage{times}
\usepackage{epsfig}
\usepackage{graphicx}
\usepackage{amsmath}
\usepackage{amssymb}
\usepackage{algorithm}
\usepackage{algpseudocode}
\usepackage{multirow}
\usepackage{diagbox}
\usepackage{authblk}

\usepackage[breaklinks=true,bookmarks=false]{hyperref}

\iccvfinalcopy 

\def\iccvPaperID{11410} 

\ificcvfinal\fi

\begin{document}

\title{Towards Understanding the Generative Capability of\\ Adversarially Robust Classifiers}

\author{





Yao Zhu$^{1}$\thanks{This work was done when he was a research intern in Huawei Noah's Ark Lab. Email: ee$\_$zhuy@zju.edu.cn.}, Jiacheng Ma$^2$, Jiacheng Sun$^{2}$\thanks{Corresponding to: Jiacheng Sun (sunjiacheng1@huawei.com)} , Zewei Chen$^2$, Rongxin Jiang$^1$, Zhenguo Li$^2$\\
\vspace{-2 mm} 
$^1$Zhejiang University,
$^2$Huawei Noah's Ark Lab

}

\maketitle
\ificcvfinal\thispagestyle{empty}\fi

\begin{abstract}
 Recently, some works found an interesting phenomenon that adversarially robust classifiers can generate good images comparable to generative models. We investigate this phenomenon from an energy perspective and provide a novel explanation. We reformulate adversarial example generation, adversarial training, and image generation in terms of an energy function. We find that adversarial training contributes to obtaining an energy function that is flat and has low energy around the real data, which is the key for generative capability. Based on our new understanding, we further propose a better adversarial training method, Joint Energy Adversarial Training (JEAT), which can generate high-quality images and achieve new state-of-the-art robustness under a wide range of attacks. The Inception Score of the images (CIFAR-10) generated by JEAT is 8.80, much better than original robust classifiers (7.50). 
\end{abstract}

\section{Introduction} \label{introduction}
Adversarial training can improve the robustness of a classifier against adversarial perturbations imperceptible to humans. Unlike a normal classifier, an adversarially robust classifier can generate good images by gradient descending the cross-entropy loss from random noises. Recently, some works discovered this phenomenon, and the quality of generated images is even comparable to GANs \cite{santurkar2019synthesis,engstrom2019prior}. 
The generative capability of an adversarially robust model from a classification task is interesting and surprising. However, it is unclear why an adversarially trained classifier can generate natural images.
As image generation is a crucial topic, understanding the generative capability of adversarially robust classifiers could be inspiring and can give hints to many other generative methods.

\begin{figure*}[ht!]
 \begin{center}  
 \centerline{\includegraphics[width=0.95\linewidth]{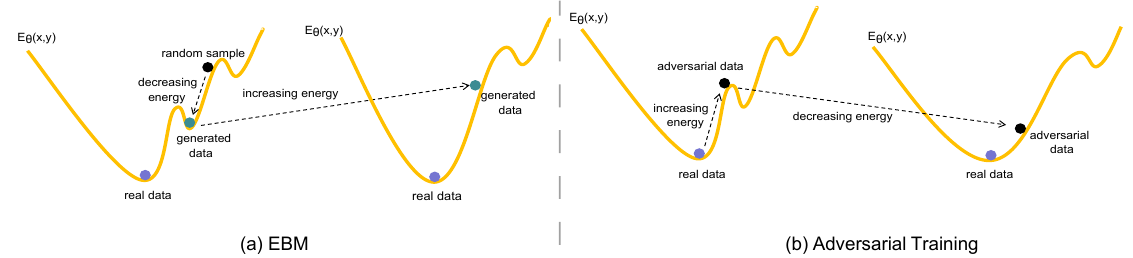}}
 \caption{EBM training and adversarial training make the energy functions smooth near the real data in different ways. (a) EBM starts from random samples, then moves along the direction of energy descent for fixed steps to obtain low energy samples. During optimization, the energy of these points is increased. Thus EBM obtains a smooth energy region around real data. (b) Adversarial examples are high-energy samples around real data. Adversarial training decreases the energy of adversarial examples by updating model parameters, thus obtains a flat region around real data.} 
 %
 \label{fig:curve_energy_all}
 \end{center}
 \vskip -0.4in
 \end{figure*}

In this paper, we aim to understand the generative capability of an adversarially trained classifier and further improve the quality of generated images from the energy perspective. For Energy-Based Model (EBM) \cite{grathwohl2020classifier}, it first generates low energy samples from random noises, then increases the energy of generated samples by updating model parameters. In this way, EBM can obtain a good energy function that is smooth and has low energy near the real data, as illustrates in Fig.~\ref{fig:curve_energy_all}(a). Thus, EBM can generate good images by sampling with Langevin Dynamics with the good energy function \cite{implicitEBM}.  

We show that adversarially trained classifiers can also obtain such a good energy function, which is flat and has low energy near the real data. This implies the generative capability of the adversarially robust classifier. For a classifier, we can define energy functions on the output logits. 
We reformulate the original adversarial training and image generation in terms of energy functions.
In fact, the adversarial examples are high-energy samples near the real data, and adversarial training is trying to decrease the energy of these examples by updating model parameters. This procedure can also help us to learn a flat energy function with low energy near the real data, as illustrated in Fig.~\ref{fig:curve_energy_all}(b). 
Moreover, based on our understanding, we find another interesting phenomenon that a normal classifier is able to generate images if we add random noise to images during training. Though injecting noises to training data can generate high-energy samples, the perturbation direction may not be efficient compared with the adversarial attack. Thus, larger noise is needed in training for generating high-energy samples.

The cross-entropy loss can be expressed as the difference between $E_{\theta}(x,y)$ and $E_{\theta}(x)$. We show that $E_{\theta}(x,y)$, which is flat and has lower energy around the real data, plays a key role in the conditional generation task. However, making the energy $E_{\theta}(x,y)$ flat and low near the real data is just a by-product of original adversarial training. 
Thus we propose Joint Energy Adversarial Training (JEAT) to directly optimize the $E_{\theta}(x,y)$. We generate adversarial examples by directly increasing the $E_{\theta}(x,y)$ as Eq. (\ref{xadv3}) and update model parameters by maximizing the likelihood of joint distribution $p_{\theta}(x,y)$ as Eq. (\ref{jeatloss}). We show that JEAT further improves the quality of the generated images.


The main contributions of our paper are summarized below:
\begin{itemize}
{\item[$\bullet$] We propose a novel explanation of the image generation capability of a robust classifier from an energy perspective.}
{\item[$\bullet$] We find the generative capability of normal classifiers with injecting noise in the training process and explain it from the energy point of view.}
{\item[$\bullet$] We propose a training algorithm JEAT from an energy perspective that improves image generative capability.
}
\end{itemize}

\begin{figure*}[t!]
\begin{center}  
\centerline{\includegraphics[width=0.95\linewidth]{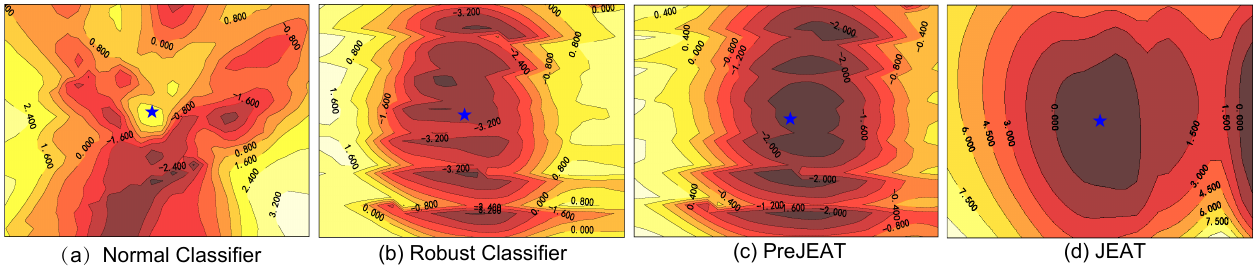}}
\caption{Energy contour of different models. (a) The energy $E_\theta(x, y)$ contour of a naturally trained model; (b) The energy contour of an adversarially trained model; (c) The energy contour of PreJEAT (ours); (d) The energy contour of JEAT (ours). Darker colors in the plots represent lower energy values. The blue star point in the center of each figure is a chosen real image from CIFAR-10. We perturb the image in two random directions to get the energy landscape.}
\label{fig:energy}
\end{center}
\vskip -0.4in
\end{figure*} 
\section{Preliminaries}

\subsection{Adversarial Training}

Adversarial training, first proposed in \cite{goodfellow2014explaining}, can effectively defend against adversarial attacks by solving a bi-level min-max optimization problem \cite{madry2017towards}. It can be formulated as:
\begin{equation}
\left\{
         \begin{array}{lr}
         \min_{\theta} \mathbb{E}_{(x,y) \sim \mathcal{D}}[\mathcal{L}(x+\delta^*,y;\theta)], &  \\
         \delta^*=\arg\max_{\left\|\delta\right\| _p\leq\epsilon}\mathcal{L}(x+\delta,y;\theta),&   \\
         \end{array}
\right.
\end{equation}
where $y$ is the ground-truth label of the input $x$, $\delta^*$ denotes the adversarial perturbation added to $x$, $\mathcal{L}$ denotes the loss function, $\left\|\cdot\right\|_p$ denotes the $\ell_p$-norm that constrains the perturbated sample in an $\ell_{p}$-ball with radius $\epsilon$ centered at $x$.

The $\ell_\infty$ adversarial perturbation is usually approximately solved by the fast gradient sign method (FGSM)~\cite{goodfellow2014explaining} and projected gradient descent (PGD)~\cite{madry2017towards}. For FGSM attacks $\delta_{x}$ takes the form:
\begin{equation}
     \delta_x = \epsilon ~  sign(\nabla_{x} \mathcal{L}(x, y;\theta)),
\end{equation}
where $sign$ is the sign function. PGD attack is a kind of iterative variant of FGSM which generates an adversarial sample starting from a random position in the neighborhood of the clean image. PGD can be formulated as:
\begin{equation}
\left\{
\begin{array}{l}   
    x_n = x_{n-1}+\eta\cdot sign(\nabla_{x_{n-1}}\mathcal{L}(x_{n-1},y;\theta)),\\
    x_n = clip(x_n,x_0-\epsilon,x_0+\epsilon),
\end{array}
\right.
\label{advpert}
\end{equation}
where $x_0$ is a clean image and $\eta$ is the perturbation step.

\subsection{Image Generation of Robust Model} \label{isrm}
Generating images with an adversarially trained classifier is first raised by Santurkar, Madry, etc. \cite{imagesynthesis} where they demonstrate that a robust classifier alone suffices to tackle various image synthesis tasks such as generation.
For a given label $y$, image generation minimizes the loss $\mathcal{L}$ of label $y$ by: 
\begin{equation}
\begin{array}{ll}
     x' = x - \frac{\eta}{2} \cdot \nabla_{x}\mathcal{L}(x, y;\theta) + \sqrt{\eta}\epsilon.
\end{array}
\label{gy}
\end{equation}
Starting from sample $x_{0} \sim \mathcal{N}(\mu_{y},\Sigma_{y})$ where $\mu_{y} = \mathbb{E}_{x\sim P_{x|y}}(x)$ and $\Sigma_{y} = \mathbb{E}_{x \sim P_{x|y}}( (x-\mu_{y})^{T}(x-\mu_{y}))$, we can obtain a better image
by minimizing the loss $\mathcal{L}$. 
Some simple improvements such as choosing a more diverse distribution to start with could further improve the quality of generated images ~\cite{imagesynthesis}. However, it is not the focus of this paper.

\subsection{Energy Based Model} \label{ebm}
Energy Based Models~\cite{lecun2006,grathwohl2020classifier} show that any probability density $p(x)$ for $x$ can be expressed as
\begin{equation}
    p_{\theta}(x) = \frac{\exp(-E_{\theta}(x))}{Z_{\theta}}, 
\label{px0}
\end{equation}   
where $E_\theta(x)$ represents the energy of $x$ and is modeled by neural network, $Z_{\theta} = \int\exp(-E_{\theta}(x))dx$ is the normalizing factor parameterized by $\theta$ also known as the partition function. The optimization of the energy-based model is through maximum likelihood learning by minimizing $\mathcal{L}_{ML}(\theta) = \mathbb{E}_{x\sim P_{D}}[ -\log p_\theta(x) ]$. The gradient is \cite{implicitEBM}:
\begin{equation}
\nabla_\theta \mathcal{L}_{ML}(\theta) = \mathbb{E}_{x^+\sim p_{D}}[ \nabla_\theta E_\theta(x^+)] - \mathbb{E}_{x^-\sim p_{\theta}}[ \nabla_\theta E_\theta(x^-)].
\label{EBMopt}
\end{equation}
Because sampling from $p_{\theta}$ is not feasible for an EBM, Langevin Dynamics is commonly used to approximately find samples from $p_{\theta}$ \cite{grathwohl2020learning}. 
Thus, EBM training usually contains two stages: approximately generating samples from $p_{\theta}$ by Langevin Dynamics along the direction of energy descent and optimizing model parameters to increase the energy of these samples and decrease the energy of real samples by SGD. In this way, as illustrated in Fig. \ref{fig:curve_energy_all} (a), EBM could obtain a smooth energy function around real data and generate samples through Langevin Dynamics.

From energy perspective, we can also define $p_\theta(x,y)$ as follows:
\begin{equation}
    p_{\theta}(x,y) =  \frac{\exp(-E_{\theta}(x,y))}{\tilde{Z_{\theta}}},
\label{pxy0}
\end{equation}
where $\tilde{Z_{\theta}} = \int\sum\limits_{y}\exp(-E_{\theta}(x,y))dx$.
Thus we also get $p_\theta(y|x)$ expressed by $E_\theta(x)$ and $E_{\theta}(x,y)$:
\begin{equation}
    p_{\theta}(y|x) = \frac{p_{\theta}(x,y)}{p_{\theta}(x)} = \frac{\exp(-E_{\theta}(x,y)) \cdot Z_{\theta}}{\exp(-E_{\theta}(x)) \cdot \tilde{Z_{\theta}}}.
\label{pxgiveny0}
\end{equation}

\subsection{Sampling with Langevin Dynamics} \label{sgldsection}
Using $\nabla_{x} \log(p(x))$, Langevin Dynamics could generate samples from density distribution $p(x)$.
The process starts from an initial point $\tilde{x}_0$ from a prior distribution $\pi$ and recursively updates $\tilde{x}$ by:
\begin{equation}
\tilde{x}_{t} = \tilde{x}_{t-1} + \frac{\eta}{2}\nabla_{\tilde{x}_{t-1}} \log (p(\tilde{x}_{t-1})) + \sqrt{\eta}\epsilon,
\label{sgld}
\end{equation}
where $\epsilon \sim \mathcal{N}(0,I)$ and $\eta$ is a fixed step size. When $\eta \rightarrow 0$ and $T\rightarrow \infty$, $\tilde{x}_T$ is exactly an sample from $p(x)$ under some condition \cite{SGLD}. If we want to sample from $p_{\theta}(x, y)$ where $y$ is a certain label, we could use Eq.~(\ref{sgld}) as well. Based on the energy framework as Eq. (\ref{pxy0}), we have $\nabla_{x} \log(p_{\theta}(x, y))=-\nabla_{x} E_{\theta}(x, y)$. Hence the sampling process becomes:
\begin{equation}
\tilde{x}_{t} = \tilde{x}_{t-1} - \frac{\eta}{2}\nabla_{\tilde{x}_{t-1}} E_{\theta}(\tilde{x}_{t-1}, y) + \sqrt{\eta}\epsilon.
\label{sgldexy}
\end{equation}

\section{Energy Perspective on Robust Classifier} \label{EPRC}

\subsection{Energy Perspective on Adversarial Training} \label{epois}
We denote $f(\cdot; \theta)$ as a classification neural network parameterized by $\theta$. Let $x$ be a sample. Then $f(x; \theta)[k]$  represents the $k^{th}$ output of the last layer and we define $p_\theta(y|x)$ as:
\begin{equation}
p_\theta(y|x) = \frac{\exp({f(x;\theta)[y]})}{\sum_{k=1}^{n} \exp({f(x; \theta)[k]})},
\label{p(ygx)}    
\end{equation}
which resembles Boltzmann distribution, and $n$ represents total possible classes. 
From Eq. (\ref{pxgiveny0}) and (\ref{p(ygx)}), we define two energy functions as follows:
\begin{equation}
\left\{
\begin{array}{l}
E_{\theta}(x,y) = -\log(\exp({f(x;\theta)[y]})), \\
E_{\theta}(x) = -\log(\sum_{k=1}^{n} \exp({f(x;\theta)[k]})). 
\end{array}
\right.
\label{energy_x}
\end{equation}
From Eq. (\ref{energy_x}), we have $\tilde{Z_{\theta}} = Z_{\theta}$ . 
Furthermore, if the classification loss function is cross-entropy loss, it could also be expressed as:
\begin{equation}
\mathcal{L}(x, y;\theta) = E_{\theta}(x,y)-E_{\theta}(x).
\label{ceenergy}    
\end{equation}

In original adversarial training as in \cite{madry2017towards,wong2020fast,shafahi2019adversarial}, by Fast Gradient Sign Method, adversarial example could be found by 
\begin{equation}
\begin{array}{ll}
     x_{adv} = x + \eta \cdot sign(\nabla_{x}(E_{\theta}(x,y)-E_{\theta}(x))),
\end{array}
\label{xadv2}
\end{equation}
which is the direction of gradient ascent of loss defined in Eq. (\ref{ceenergy}). The direction of adversarial perturbation relates to not only $E_{\theta}(x,y)$ but also $E_{\theta}(x)$.
And in original adversarial training, the optimization process aims to decrease the loss in Eq.~(\ref{ceenergy}) by updating model parameters. 


\begin{figure*}[ht!]
 \begin{center}  
 \centerline{\includegraphics[width=0.95\linewidth]{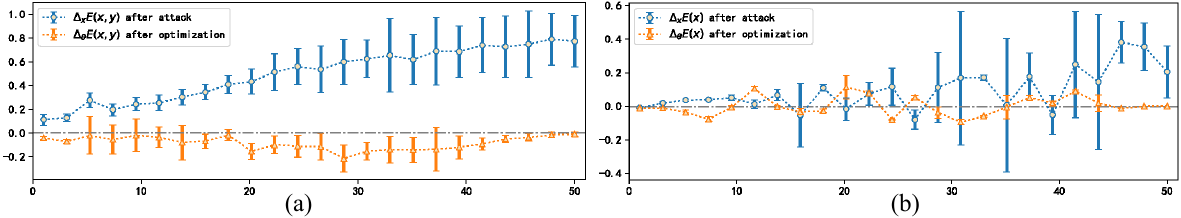}}
 \caption{We illustrate the changes of energy in original adversarial training \cite{madry2017towards} on CIFAR-10 in 50 epochs (model has converged). 
 The center points of the tags represent the mean value and the lengths represent the variance.
 (a) Adversarial examples increase the energy $E_\theta(x,y)$ as $\Delta_x E_{\theta}(x,y)$\textgreater 0 during the training. The energy $E_\theta(x,y)$ of adversarial examples decrease after updating parameters as $\Delta_\theta E_{\theta}(x,y)$\textless 0 during the training. (b) The value of $\Delta_\theta E_\theta(x)$ fluctuates around zero, sometimes positive and sometimes negative. Thus $E_\theta(x)$ has not been well optimized in the classification task.}
 \label{fig:delta_energy}
 \end{center}
 \vskip -0.4in
 \end{figure*}


As we mentioned in Sec. \ref{ebm}, a good energy function $E_{\theta}(x,y)$ which is smooth and has low energy around real data is the key factor for generating good images for a given label $y$. 
We define the change of energy $E_{\theta}(x,y)$ after adversarial attack and after optimization as:
\begin{equation}
\left\{
\begin{array}{l}
\Delta_x E_{\theta}(x,y) = E_{\theta}(x_{adv},y)-E_{\theta}(x_{ori},y),\\
\Delta_\theta E_{\theta}(x,y) = E_{\theta_{updated}}(x_{adv},y)-E_{\theta}(x_{adv},y),
\end{array}
\label{deltaE}
\right.
\end{equation}
where $E_{\theta}(x_{adv},y)$ is the energy of $x_{adv}$ given label $y$ before updating parameters, $E_{\theta}(x_{ori},y)$ is the energy of $x_{ori}$ given label $y$ before updating parameters, and $E_{\theta_{updated}}(x_{adv},y)$ is the energy of $x_{adv}$ given label $y$ after updating parameters. The definition of $\Delta_x E_{\theta}(x)$ and $\Delta_\theta E_{\theta}(x)$ are similar in (\ref{deltaE}) with removing $y$.

As illustrated in Fig. \ref{fig:delta_energy} (a), adversarial attacks generate high-energy adversarial examples, and then the energy of adversarial examples is decreased by updating model parameters.
Compared to EBM, adversarial training flattens the energy region around real data in a way different from EBM training. Adversarial training finds adversarial examples with high energy $E_\theta(x,y)$ near the data and then decreases the energy of those samples by updating model parameters. Noted that, in EBM, sampling from $p_{\theta}(x,y)$ by Langevin Dynamics follows the direction of energy descent, which is the opposite direction of adversarial examples. As illustrated in Fig. \ref{fig:curve_energy_all} (b), adversarial training can obtain a flat energy function around real data. 

We illustrate energy $E_\theta(x, y)$ contours in Fig.~\ref{fig:energy} for a normal classifier and an adversarially trained classifier. 
From Fig.~\ref{fig:energy} (a), the energy function around the center is sharp for a normal classifier, and the energy of the real data is high. Because the low-energy region deviates from the center in a normal classifier, the generated images along the direction of energy descent are likely to fall into a region far away from the real data, which would not have good quality. For a robust classifier, the energy function near the center is smoother and lower, as shown in Fig.~\ref{fig:energy} (b).

Image generation by a robust classifier is introduced in Sec.~\ref{isrm}, if we transform the loss function to energy expression as Eq. (\ref{ceenergy}), the generating procedure of images becomes:
\begin{equation}
\begin{array}{ll}
     x' = x - \frac{\eta}{2} \cdot \nabla_{x}(E_{\theta}(x,y)-E_{\theta}(x)) + \sqrt{\eta}\epsilon,
\end{array}
\label{gimage}
\end{equation}
where $\epsilon \sim \mathcal{N}(0, I)$ and $\eta$ is step size.
The term $E_{\theta}(x,y)-E_{\theta}(x)$ in Eq. (\ref{gimage}) implies that the generation iterations are related to both the energy of $E_{\theta}(x,y)$ and the energy of $E_{\theta}(x)$. By exploring low energy region of  $E_{\theta}(x,y)$, which corresponds to high probability $p_{\theta}(x,y)$ region given constant normalizing factor from Eq. (\ref{pxy0}), we could obtain a good sample from high probability $p_{\theta}(x,y)$ region. 
Minimizing $E_{\theta}(x,y)$ contributes to maximizing $p_{\theta}(x,y)$ to generate images corresponding to the label $y$ as shown in Eq.~(\ref{sgldexy}), but the energy of $E_{\theta}(x)$ is irrelevant to label $y$. 



\begin{figure}[ht!]
 \begin{center}  
 \centerline{\includegraphics[width=1.0\linewidth]{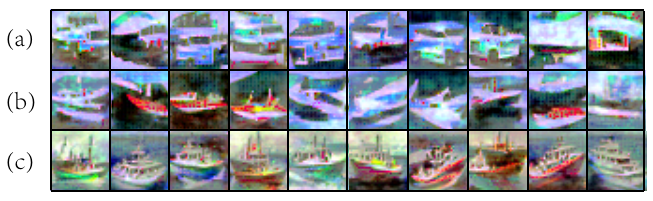}}
 \caption{Given the label ship, the images generated by different methods. (a) shows ten images generated by original robust classifier using Eq.~(\ref{gimage}). (b) shows ten images generated by original robust classifier using Eq.~(\ref{gimage2}). (c) shows ten images generated by robust classifier trained with adversarial examples got by Eq.~(\ref{xadv3}) using Eq.~(\ref{gimage2}).}
 \label{fig:ex_exy}
 \end{center}
 \vskip -0.35in
 \end{figure}

As illustrated in Fig. \ref{fig:delta_energy} (b), $E_{\theta}(x)$ has not been well optimized in the classification task, which may introduce label-independent noise. 
Thus, $E_{\theta}(x)$ may be a factor restricting the generative capability of the robust classifier,
and we could drop $E_{\theta}(x)$ while only using $E_{\theta}(x,y)$ as: 
\begin{equation}
     x' = x - \frac{\eta}{2} \cdot \nabla_{x}(E_{\theta}(x,y)) + \sqrt{\eta}\epsilon.
\label{gimage2}
\end{equation}
Eq.~(\ref{gimage2}) is closely relate to sampling from Langevin Dynamics as Eq.~(\ref{sgldexy}). Approaching low energy region of $E_{\theta}(x,y)$ is equivalent to approaching the high probability region of $p_{\theta}(x, y)$. 
As shown in Fig.~\ref{fig:ex_exy}(a) and (b), using Eq.~(\ref{gimage2}) gives better generated images than using Eq.~(\ref{gimage}). 

 \subsection{Endowing Generative Capability to Normal Classifier}
 
 \begin{figure}[htbp!]
 \begin{center}  
 \centerline{\includegraphics[width=1.0\linewidth]{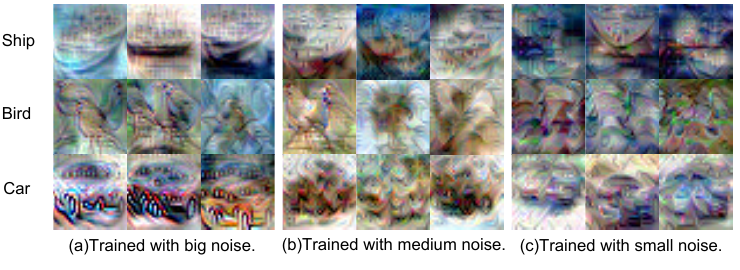}}
 \caption{The images generated by the models injected different strengths of noise in the training process. The clean accuracy of these models is almost the same. We use $\delta$ to denote the noise added to training data. (a) $\delta \sim \mathcal{N}(0,24/255)$. Obviously, this model produces recognizable images. (b) $\delta \sim \mathcal{N}(0,16/255)$, the shape of ship is visible. (c) $\delta \sim \mathcal{N}(0,8/255)$, it is hard to recognize the shape of objects.}
 \label{fig:noise}
 \end{center}
 \vskip -0.4in
 \end{figure}
 As analyzed before, we can generate high-energy samples from real data and reduce the energy of these samples during the optimization process to obtain a classifier with generative capability. The process of optimizing the energy $E_{\theta}(x,y)$ to be flat and low around real data contributes to generating good images. Adversarial training follows the procedure and provides an effective way to find high-energy samples with small perturbations. 
 Meanwhile, a simple random noise could also find high-energy samples but less effectively. 
 
 To verify our claim, we train normal classifiers by injecting different strengths of random noise into training data. 
As shown in Fig.~\ref{fig:noise}, classifiers trained with different strengths of noise generate images of different qualities through the procedure in Sec. \ref{epois}. Though the quality is not good enough, cars, birds and ships can be recognized with high confidence in Fig.~\ref{fig:noise} (a). From the comparison of images generated by adding different strengths of noise, we find that stronger random noise is needed to train a normal classifier in order to generate recognizable images. This is consistent with our explanation above that random noise is a ``blind" adversarial direction to find high-energy samples. Because the direction is uncertain and very likely not the direction to efficiently increase energy, stronger perturbations may be needed. 
This experiment also shows that the process which generates high-energy samples from real data and reduces the energy of these samples during the optimization contributes to obtaining a classifier with generative capability.
 
 

\subsection{Generating Adversarial Examples by Energy} \label{epoae}


As mentioned above, $E_\theta(x,y)$ determines the generative capability of a robust classifier. In original adversarial training, it does not increase the energy $E_\theta(x,y)$ of original data directly.
In fact, only using $E_{\theta}(x,y)$ can also make $x$ to be an adversarial sample $x_{adv}$ as shown in Tab. \ref{energy-attack-table}. 


\begin{table}[ht]
	\caption{Normal attack and energy attack which just increases $E_{\theta}(x,y)$ on normal classifier (WideResNet-28-10). The perturbation radius is 8/255. They perform similarly on the CIFAR-10 and CIFAR-100 datasets and they can successfully attack the classifier.}
	\label{energy-attack-table}  
	\begin{center}
	\begin{tabular}{c|c|c}  
		\hline
		Attack&CIFAR-10&CIFAR-100\\
		\hline
		No Attack&94.39$\%$&78.54$\%$\\
		\hline
		Normal Attack&0.12$\%$&0.08$\%$\\
		\hline
		Energy Attack&0.11$\%$&0.10$\%$\\
		\hline		
	\end{tabular}
	\end{center}
    \vskip -0.25in
\end{table}
Using $E_{\theta}(x,y)$ only in Eq. (\ref{xadv2}), we have:
\begin{equation}
     x_{adv} = x + \eta \cdot sign(\nabla_{x}(E_{\theta}(x,y))).
\label{xadv3}
\end{equation}
Adversarial examples generated by energy attack in Eq.~(\ref{xadv3}) have almost the same attack effect with the original normal attack by Eq. (\ref{xadv2}). The difference between $x$ and $x_{adv}$ is unrecognizable, but their energy are quite different. We name the method which trained with adversarial examples found by Eq.~(\ref{xadv3}) and generates images by Eq.~(\ref{gimage2}) after training as Preliminary Joint Energy Adversarial Training (PreJEAT).

As we illustrate in Fig.~\ref{fig:energy}, in a PreJEAT trained classifier, the energy around the natural image (center) is flatter and lower than a normal classifier and original robust classifier. In a robust classifier, the low energy region deviates from the center. But in the PreJEAT trained classifier, the natural image has the lowest energy. Thus, directly generating adversarial examples by $E_{\theta}(x,y)$ contributes to obtaining a flat and low energy function near the real data.
By adversarial training with Eq.~(\ref{xadv3}) we can obtain a robust classifier with better generative capability, as shown in Fig.~\ref{fig:ex_exy}(c). 

\begin{algorithm}[t]
    \caption{Training and Generating of JEAT: Given network $f$,  $E(x,y)$ = $-\log(\exp({f(x)[y]}))$ represent energy of $(x,y)$,  $E(x)$ = $-\log(\sum_{y} \exp({f(x)[y]}))$ represent energy of $x$, $\ell_{\infty}$ adversarial perturbation radius $\epsilon$, SGLD step-size $\alpha$, SGLD steps $K$, replay buffer $B$, reinitialization probility $\rho$, epochs $T$, dataset of size $M$, learning rate $\eta$.}
    \begin{algorithmic}
    \State {\textbf{Training}}: 
    \For {$i = 1,2...,T$}
        \For {$j = 1,2...,M$} 
            \State $\blacktriangleright$ Generating energy-based adversarial samples:
            \State $\delta=\mathcal{U}(-\epsilon,\epsilon)$
            \State $\delta=\delta+\epsilon \cdot sign(\nabla_{x}(E_{\theta}(x,y)))$
            \State $\delta=max(min(\delta,\epsilon),-\epsilon)$
            \State $x_{adv}=x_j+\delta$

            \State $\blacktriangleright$ Generating samples by SGLD:
            \State $\tilde{x}_0 \sim \mathcal{U}(0,1)$ with probability  $\rho$, else $\tilde{x}_0 \sim B$
            \For {$t = 0,1,2,...,K-1$}
                \State
                $\tilde{x}_{t+1} = \tilde{x}_{t} - \frac{\alpha}{2}\cdot\nabla_{x_{t}}E_{\theta}(\tilde{x}_{t}) + \sqrt{\alpha} \cdot \mathcal{N}(0,I)$
		    \EndFor 
            \State Add $\tilde{x}_{K}$ to $B$
            \State $\blacktriangleright$ Updating model parameters:
            \State $\nabla_{\theta}\mathcal{L}_{p(y|x_{adv})} =\nabla_{\theta} (E_{\theta}(x_{adv},y) - E_{\theta}(x_{adv}))$
            \State $\nabla_{\theta}\mathcal{L}_{p(x_{adv})}=\nabla_{\theta} (E_{\theta}(x_{adv}) - E_{\theta}(\tilde{x}_{K}))$
            \State $\theta=\theta-\eta \cdot (\nabla_{\theta}\mathcal{L}_{p(y|x_{adv})}+\nabla_{\theta}\mathcal{L}_{p(x_{adv})})$
        \EndFor
	\EndFor

  \State
  \State
    {\textbf{Generating}}:
        \State $x_{0}$ $\sim$ \text{random sample}
        \For {$t = 0,1,2,...,K-1$}
            \State
            $x_{t+1} = x_{t} - \frac{\alpha}{2}\cdot\nabla_{x_{t}}E_{\theta}(x_{t},y) + \sqrt{\alpha} \cdot \mathcal{N}(0,I)$
		\EndFor
		\State {\textbf{Output}:} $x_{gen}=x_{K}$
    \end{algorithmic}
    \label{algorithm:JEAT-algorithm}
\end{algorithm}

\subsection{Joint Energy Adversarial Training}\label{JEAT}


We verified that a PreJEAT trained model has good generative capability in the previous section. However, there is still a discrepancy between training loss objective as Eq. (\ref{ceenergy}) and adversarial training as Eq. (\ref{xadv3}) in PreJEAT. And we also find that the images generated by PreJEAT are not smooth enough in Fig.~\ref{fig:ex_exy}(c). We hope that the optimization process also optimizes $E_{\theta}(x,y)$ more directly to reduce the energy around real data. Hence we propose a new algorithm, Joint Energy Adversarial Training (JEAT), in Algorithm \ref{algorithm:JEAT-algorithm} to improve the generative capability of a classifier. In JEAT, we replace cross-entropy loss $-\log p_\theta(y|x)$ in PreJEAT with :
\begin{equation}
-\log p_\theta(x,y) =-\log p_\theta(y|x)-\log p_\theta(x), \label{jeatloss}
\end{equation}
where $p_\theta(x)$ is defined in Eq. (\ref{px0}). Optimizing $p_{\theta}(x,y)$ helps to get better $E_{\theta}(x,y)$ as shown in Eq. (\ref{pxy0}). The gradient of $\log p_\theta(x)$ is 
\begin{equation}
    \begin{array}{ll}
    \nabla_{\theta}{\log(p_{\theta}(x))} = - \nabla_{\theta}{E_{\theta}(x)} - \mathbb{E}_{p_{\theta}(x)}[\nabla_{\theta}{E_{\theta}(x)}].
    \end{array}
    \label{gpx}
\end{equation}
We use Stochastic Gradient Langevin Dynamics (SGLD) to approximate $\mathbb{E}_{p_{\theta}(x)}$~\cite{du2020implicit}.

 JEAT uses Eq.~(\ref{xadv3}) to find adversarial example and Eq.~(\ref{gimage2}) to generate image like PreJEAT.
 With our proposed JEAT, adversarial examples, adversarial training, and image generation are connected to the energy $E_\theta(x,y)$ in a clear way. 
We also plot the energy contour of $E_\theta(x, y)$ in Fig.~\ref{fig:energy}(d). Compared to the other three models, the energy function of a JEAT trained classifier near the center is the flattest among the four classifiers. Moreover, the energy contour is smooth across different energy levels. Hence images generated from a JEAT trained classifier are more likely a natural image and exhibits many natural details. Moreover, a flat energy function is less sensitive to noise perturbation. We will verify in experiments that JEAT improves both quality of generated images and adversarial robustness systematically.

\begin{figure*}[htb]
\begin{center}  
\centerline{\includegraphics[width=1.0\linewidth]{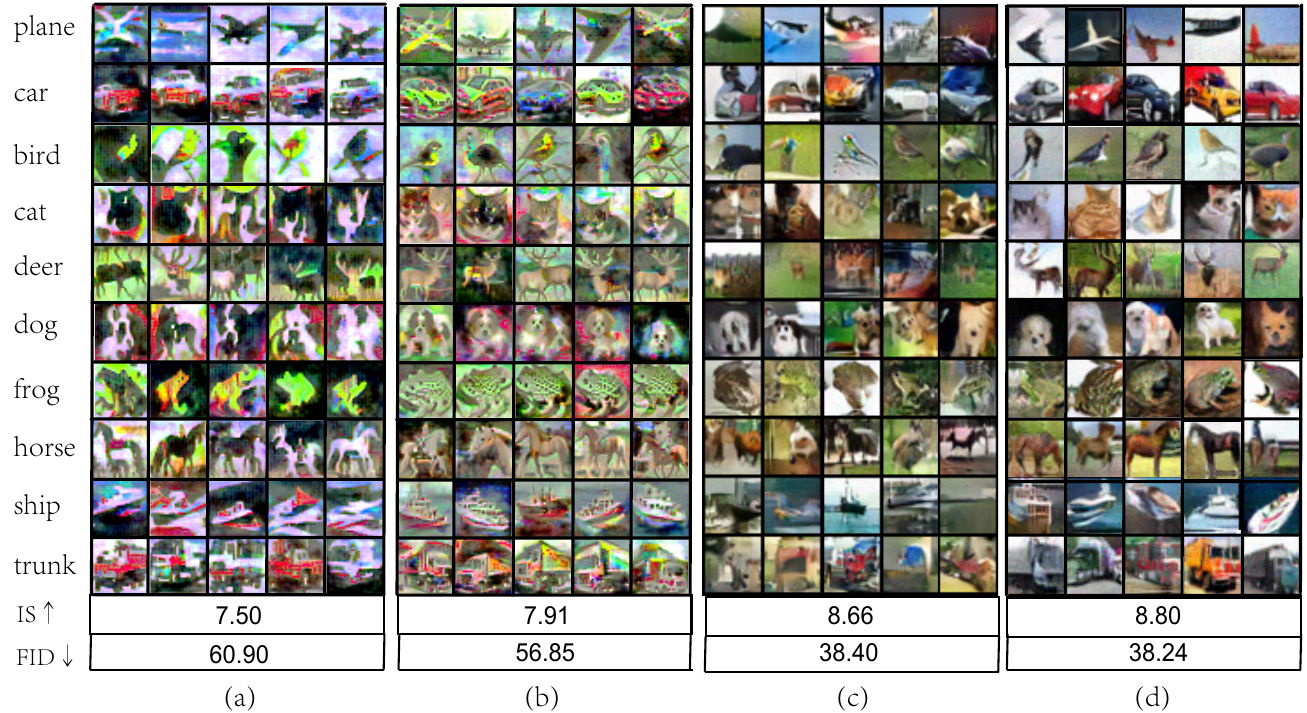}}
\caption{The generated images of different models. (a) The images generated by standard adversarially trained classifier; (b) The images generated by PreJEAT trained classifier; (c) The images generated by JEM \cite{grathwohl2020classifier} trained classifier; (d) The images generated by JEAT trained classifier. Images in each line are generated for a class in CIFAR-10.}
\label{fig:generate}
\end{center}
\vskip -0.4in
\end{figure*}

\section{Experiments}
\subsection{Experimental settings}\label{setting}
To verify the validity of our energy-based explanation and the effectiveness of JEAT, we conducted experiments on CIFAR-10, CIFAR-100 and compare them with other methods. For network architecture, we use the WideResNet-28-10 for all algorithms. We repeat our experiments five times and report the mean and standard deviation.

In terms of image generation, we compare our generated pictures with JEM~\cite{grathwohl2020classifier} and robust classifier~\cite{imagesynthesis}. We show the images generated by different algorithms and the scores of these methods based on Inception Score and Frechet Inception Distance metrics.


\subsection{Image Generation} \label{experiment}
We use the procedure described in Sec. \ref{epois} to generate images by using different classifiers. The images generated by the adversarially trained classifier are shown in Fig.~\ref{fig:generate}(a). 
%
%
As analyzed in Sec. \ref{epois}, $E_\theta(x)$ can be the factor that restricts the generative capability of a robust model. We show the images generated by PreJEAT which uses $E_\theta(x,y)$ instead of $E_\theta(x,y)-E_\theta(x)$ both in adversarial training and generating images in Fig.~\ref{fig:generate}(b). This approach greatly improves the quality of generated images. The energy plots of classifiers are presented in Sec. \ref{EPRC}. By using PreJEAT, the energy is smoother and has a larger low energy region around real data, and it generates better quality images, especially in fine details and overall shape.


 We also show the images generated by JEM~\cite{grathwohl2020classifier} which is an energy-based model utilizing a classifier in Fig.~\ref{fig:generate}(c). Compared to Fig.~\ref{fig:generate}(b), JEM generates a more diverse background and is smooth in pixel. However, the features of objects stand out and are more perceptible to human in the images generated by PreJEAT.  
 
  \begin{figure}[bt]
 \begin{center}

 \centerline{\includegraphics[width=1.0\linewidth]{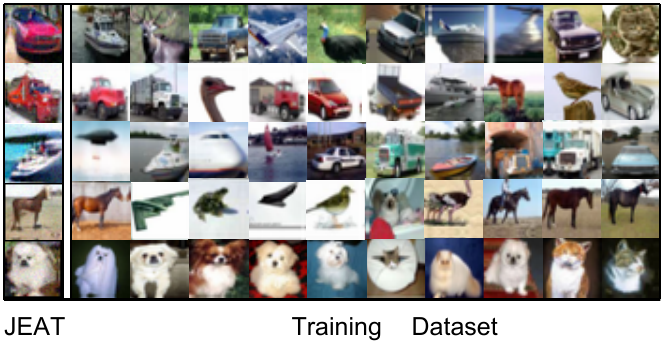}}
 \caption{JEAT generates different images different from training data. The left column is generated by JEAT, and the right ten columns are images in the training dataset with the minimum distance and highest similarity to the image generated by JEAT. The image we generated is different from the original ones in the training dataset. We use SSIM~\cite{SSIM} to measure similarity.}
 \label{fig:generalization}
 \end{center}
 \vskip -0.4in
 \end{figure}
 
 As is shown in Fig.~\ref{fig:generate}(d), our JEAT algorithm generates the best images compared to the other three. The reason behind is that JEAT is adversarially trained with energy $E_{\theta}(x,y)$ as in Eq. (\ref{xadv3}), and directly optimize $E_\theta(x,y)$ in the training contributes to getting better $E_\theta(x,y)$. 
 The classifier trained by JEAT generates natural images with abundant features. Moreover, JEAT has the best performance based on the metrics of the Inception Score and Frechet Inception Distance.
 
 We also validate the generalizability of the JEAT trained classifier in the sense that the images generated differ from any images in the training set, as shown in Fig.~\ref{fig:generalization}. Hence, JEAT is not just memorizing training images it has seen but generalizes as well. 


\subsection{Score Matching Metric}
``Score" is defined here as $s_{\theta}(x, y)=\nabla_{x} \log p_{\theta}(x, y)$ \cite{score-based-gi}. 
Score matching tries to match the vector field of the gradient of $\log p_{\theta}(x,y)$ with respect to $x$ to the real vector field given in data distribution \cite{score-based-gi} by minimizing Fisher Divergence:
\begin{equation}
\begin{array}{ll}
\frac{1}{2} \cdot \mathbb{E}_{p_{data}(x|y)} [\left\|\nabla_{x} \log p_{\theta}(x,y)- \nabla_{x}\log p_{data}(x,y)\right\| _{2}^{2}].
\end{array}
\label{sm}   
\end{equation}
With the score defined in the distribution of $(x,y)$ as $s_{\theta}(x,y)$, an equivalent optimization problem to (\ref{sm}) is minimizing 
 \begin{equation}
\begin{array}{ll}
\mathbb{E}_{p_{data}(x|y)} [ \frac{1}{2} \cdot \left\|s_{\theta}(x,y)\right\| _{2}^{2} + tr(\nabla_{x}  s_{\theta}(x, y))],
\end{array}
\label{sme}   
\end{equation}
 where $\nabla_{x}  s_{\theta}(x, y)$ denotes the Jacobian of $s_{\theta}(x, y)$ \cite{JMLR:v6:hyvarinen05a}. 
From energy perspective, score $s_{\theta}(x,y)=-\nabla_{x}E_{\theta}(x,y)$.
Hence we denote objective in Eq. (\ref{sme}) as Joint Fisher Score (FS):
\begin{equation}
\begin{array}{ll}
\mathbb{E}_{p_{data}(x|y)} [ \frac{1}{2} \cdot \left\|\nabla_xE_{\theta}(x,y)\right\| _{2}^{2} - tr(\nabla_{x}^2  E_{\theta}(x, y))].
\end{array}
\label{smeenergy}   
\end{equation}
It is a simple and effective metric that a lower value indicates $\nabla_{x}\log p_{\theta}(x,y)$ is closer to $\nabla_{x} \log p_{data}(x,y)$. 
So smaller value of Fisher Score corresponds to better generative capability (not considering failure cases here). Fisher Score could be used as a metric to assess the generative capability of models. 
Similarly, we denote Marginal Fisher Score (MFS) as:
\begin{equation}
\begin{array}{ll}
\mathbb{E}_{p_{data}(x)} [ \frac{1}{2} \cdot \left\|\nabla_xE_{\theta}(x)\right\| _{2}^{2} - tr(\nabla_{x}^2  E_{\theta}(x))].
\end{array}
\label{smeenergy2}   
\end{equation}
As shown in Tab.~\ref{VFMS}, both the Joint Fisher Score and the Marginal Fisher Score for our JEAT are the smallest in all the four models, which implies that the score of our model matches well with the ground truth distribution. This also means that JEAT can generate better images using Langevin Dynamics. The score matching metric is also consistent with the results of IS and FID, which are commonly used.
\begin{table}[ht]
	\caption{Metrics for different models. ``JFS" and ``MFS" are the scores to evaluate the Joint and marginal Fisher Scores. 
	``Normal" denotes normal training, ``AT" denotes adversarial training.}
	\label{VFMS}  
	\begin{center}
	\begin{tabular}{c|c|c|c|c}  
		\hline
		Method&Normal &AT \cite{imagesynthesis}&JEM \cite{grathwohl2020classifier}&JEAT\\
		\hline
		JFS$\downarrow$&6.97&2.83&0.330&0.023\\
		\hline
		MFS$\downarrow$&23.08&3.12&1.44&0.15\\		
		\hline
		IS$\uparrow$&-&7.50&8.66&8.80\\
		\hline
		FID$\uparrow$&-&60.90&38.40&38.24\\
		\hline
	\end{tabular}
	\end{center}
    \vskip -0.2in
\end{table}

\subsection{Robustness of JEAT}

 In this section, we show that the JEAT models not only can generate good images but also has comparable robustness with other hybrid models. 
 We compare our model among hybrid models, including Glow \cite{kingma2018glow}, IGEBM \cite{IGEBM}, and JEM \cite{grathwohl2020classifier}. As shown in Tab. \ref{Robusttable}, we compare standard classification accuracy, and robust accuracy under $\ell_{\infty}$ PGD-20 attack with $\epsilon=8/255$. The experimental results show that JEAT can both improve the generative capability and adversarial robustness of JEM. 
 
 JEM \cite{grathwohl2020classifier} is a well-written paper, which proposes that classifiers can also be trained as generative models and such models have the robustness comparable to adversarial training. 
 However, when we use standard method \cite{benchmark} for evaluation, JEM's robustness is only 6.11$\%$ (under $\ell_{\infty}$ PGD-20, $\epsilon$=8/255). We follow the same standard evaluation \cite{benchmark} method to test our JEAT's robustness and show it can improve from 6.11\% to 30.55\% (under $\ell_{\infty}$ PGD-20, $\epsilon$=8/255).
\footnotetext[1]{We test the robustness of JEM's~\cite{grathwohl2020classifier} open-source model (from https://github.com/wgrathwohl/JEM) by standard evaluation method~\cite{benchmark} (https://robustbench.github.io/).
\cite{ATJEBM} also shows that the robustness of JEM is 9.29$\%$ under the same setting ($\ell_{\infty}$ PGD-20, $\epsilon$=8/255) by their implementation.}
 \begin{table}[ht]
	\caption{The performance comparison among four hybrid models on CIFAR-10. We use $\ell_{\infty}$ PGD-20 attack with $\epsilon=8/255$.}
	\label{Robusttable}  
	\begin{center}
	\begin{tabular}{c|c|c}  
		\hline
	    Model&Standard Accuracy&Robustness\\
		\hline
		Glow \cite{kingma2018glow}&67.6\%&-\\
		\hline
		IGEBM \cite{IGEBM}&45.06\%&32.19\%\\
		\hline
		JEM \cite{grathwohl2020classifier}&92.90\%&6.11\%\footnotemark[1]\\
		\hline
		JEAT&85.16\%&30.55\%\\
		\hline
	\end{tabular}
	\end{center}
\end{table}

 

\section{Related Work}
\subsection{Adversarial Training}

Adversarial attacks can give the wrong prediction while adding negligible perturbations for humans to input data~\cite{szegedy2013intriguing}. Adversarial training first proposed in~\cite{goodfellow2014explaining} can effectively defend against such attacks by training on both clean data and adversarial examples.
\cite{madry2017towards} formulates adversarial training as a bi-level min-max optimization problem and trains models exclusively on adversarial images rather than both clean and adversarial images. 
Although it effectively improves the adversarial robustness, 
expensive computational cost and performance degradation on clean images are the two fatal shortcomings of it. 
Many works try to reduce the computation cost to the natural training of a model. 
\cite{shafahi2019adversarial} updates the perturbation and weights at the same time, which reduces the bi-level problem to single-level optimization. 
\cite{wong2020fast} samples the perturbation randomly in every iteration.
Many works try to mitigate the performance degradation on clean images. 
\cite{tradesoff} gives a hint on how to balance the trade-off between vanilla and robust accuracy.
\cite{gowal2020uncovering} shows that additional unlabeled data can help to increase accuracy in adversarial training.
Interestingly, robust optimization can be recast as a tool for enforcing priors on the features learned by deep neural networks \cite{engstrom2019prior}. 
Moreover, a robust classifier can tackle some challenging tasks in image synthesis \cite{santurkar2019image}. 

\subsection{Energy-based Model}
Recently, the energy-based model has attracted significant attention. Effective estimating and sampling the partition function is the primary difficulty in training energy-based models \cite{lecun2006,hoffman2019neutralizing,wang2019implicit}.
Some research works have made a contribution to improve the training of energy-based models, such as sample the partition function through amortized generation \cite{nijkamp2019learning,dai2020exponential}, utilize a separate generator network for negative image sample generations \cite{kumar2019maximum,song2020generative} and score matching where the gradients of an energy function are
trained to match the gradients of real data \cite{song2020distribution,saremi2018deep}.
Kevin Swersky et~al. propose that treat classifier as an energy-based model can enable classifiers to generate samples rivaling the quality of recent GAN approaches \cite{grathwohl2020classifier}.
Kyungmin et~al. present a hybrid model which is built upon adversarial training and energy based training to deal with both out-of-distribution and adversarial examples \cite{ATJEBM}. 

Motivated by these discoveries, we investigate the generative capability of an adversarially trained model from an energy perspective and provide a novel explanation. We have further proposed an algorithm that can improve generation capacity.

\section{Conclusion}
 We present a novel energy perspective on the generative capability of an adversarially trained classifier and propose our JEAT methods to obtain a classifier with stronger robustness that generates images with better quality. We find that a normal classifier can also generate images by injecting random noise in the training process, and we interpret this as blind adversarial training, for that random noise may find high energy examples accidentally. 
 In summary, adversarial examples, blind or not, aim to find high-energy examples, and the classifier's optimization aims to lower the energy by optimizing model parameters. This process, as we validate, is beneficial for the robustness of the model and the quality of generated images. 
 In addition, larger model capacity \cite{gowal2020uncovering, madry2017towards}, smoother activation function \cite{gowal2020uncovering, xie2020smooth} and unlabeled data \cite{gowal2020uncovering, carmon2019unlabeled} are shown to improve adversarial robustness as well. We think these methods are promising ways to boost the performance of JEAT further and leave them for future work.

{\small
\bibliographystyle{ieee_fullname}
\bibliography{egbib}
}

\clearpage
\appendix
\section{Generated Images on CIFAR-100}
\begin{figure}[ht]
\vskip 0.2in
\begin{center}  
\centerline{\includegraphics[width=1.0\linewidth]{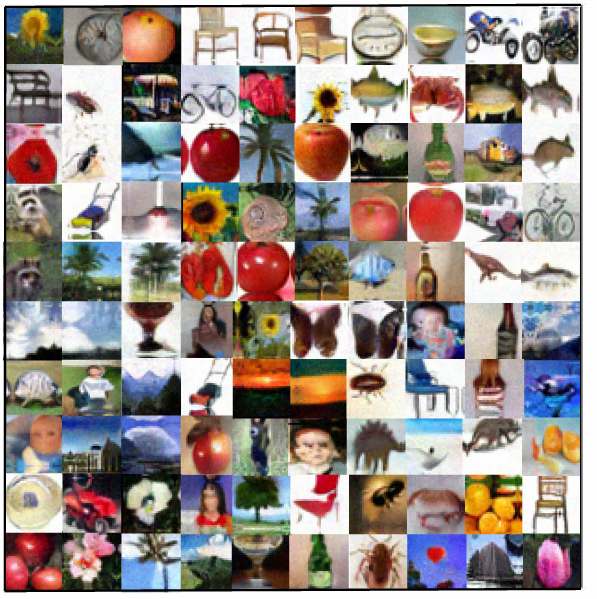}}
\caption{Images generated by JEAT on CIFAR-100.}
\label{fig:cifar100}
\end{center}
\vskip -0.2in
\end{figure}

We show the images generated by JEAT on CIFAR-100 in Fig \ref{fig:cifar100}. These images are rich in details and vivid.

\begin{table*}[ht]
	\centering
	\caption{Hyperparameters of different methods.} 
    \vskip 0.1in   
    \begin{center}
    \begin{small}
    \begin{sc}
	\begin{tabular}{|c|c|c|c|c|c|c|c}  
		\hline
		Hyperparameter& Learning Rate &Batch Size&SGD Scheduler&Epochs&Weight Decay&Momentum\\
		\hline
		JEAT(ours)&1e-4&64&Adam[160,180]&200&0&beta=[0.9,0.999]\\
		JEM&1e-4&64&Adam[160,180]&200&0&beta=[0.9,0.999]\\
		Free&0.1&128&MultiStep[100,150]&200&2e-4&0.9\\
		Fast&0-0.2&128&Cyclic&15&5e-4&0.9\\
		TRADES&0.1&128&MultiStep[75,90,100]&120&2e-4&0.9\\
		\hline
	\end{tabular}
	\end{sc}
    \end{small}
    \end{center}
	\label{hyperparameters}
\vskip -0.1in
\end{table*}

\section{The Change of Energy}
\subsection{The Change of Energy $E_\theta(x,y)$ in Training}
\begin{figure}[ht!]
 \vskip 0.2in
 \begin{center}  
 \centerline{\includegraphics[width=1.0\linewidth]{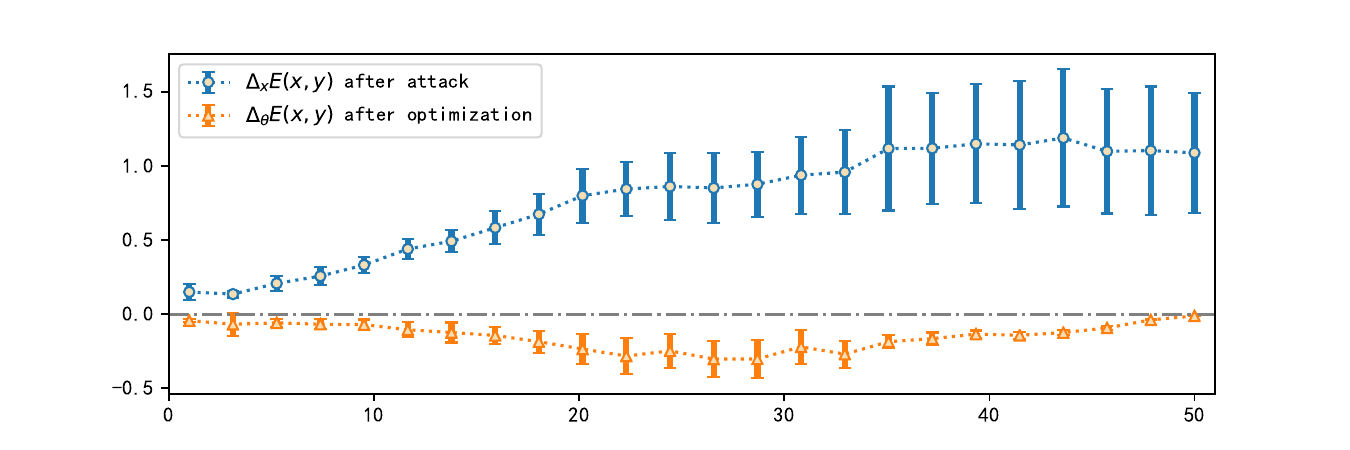}}
 \caption{We illustrate the changes of energy in original adversarial training \cite{madry2017towards} on CIFAR-100 in 50 epochs (model has converged). The center points of the tags represent the mean value and the lengths represent the variance. Adversarial examples increase the energy $E_\theta(x,y)$ as $\Delta_x E_{\theta}(x,y)>0$ during the training. The energy $E_\theta(x,y)$ of adversarial examples decrease after updating parameters as $\Delta_\theta E_{\theta}(x,y)<0$ during the training.}
 \label{fig:delta_energy2}
 \end{center}
 \vskip -0.2in
 \end{figure}

We illustrate the change of energy $E_\theta(x,y)$ in adversarial training on CIFAR-100 in Fig. \ref{fig:delta_energy2}. Adversarial attacks generate high-energy adversarial examples, and then the energy of adversarial examples is decreased by updating model parameters.
As we show in our paper, adversarial training flattens the energy region around real data in this way. 

\subsection{The Change of Energy $E_\theta(x)$ in Training}
\begin{figure}[ht!]
 \vskip 0.2in
 \begin{center}  
 \centerline{\includegraphics[width=1.0\linewidth]{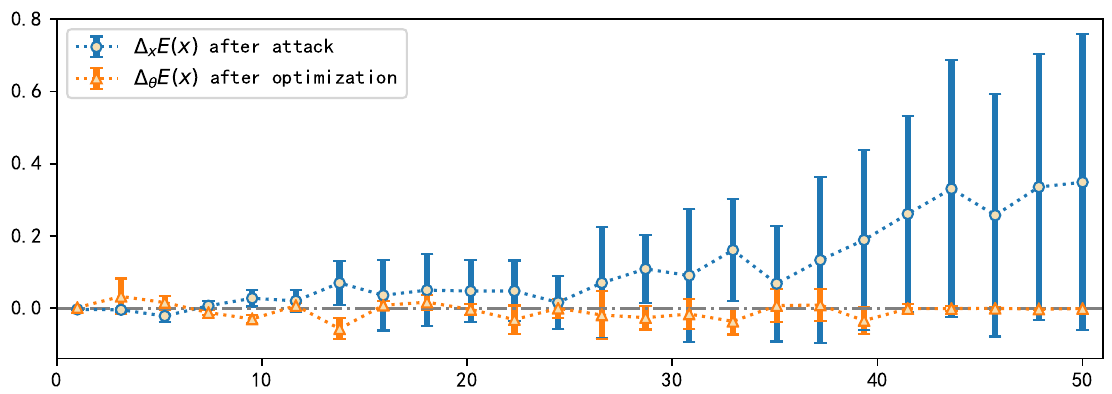}}
 \caption{We illustrate the changes of energy $E_\theta (x)$ in original adversarial training \cite{madry2017towards} on CIFAR-100 in 50 epochs (model has converged). The center points of the tags represent the mean value and the lengths represent the variance.}
 \label{fig:delta_energy_ex_cifar100}
 \end{center}
 \vskip -0.2in
 \end{figure}

We illustrate the change of energy $E_\theta(x)$ in adversarial training on 
CIFAR-100 in Fig. \ref{fig:delta_energy_ex_cifar100}. The value of $\Delta_\theta E_\theta(x)$ fluctuates around zero, sometimes positive and sometimes negative. Thus $E_\theta(x)$ has not been well optimized in the classification task. Using $E_\theta(x)$ in the generation task by a robust classifier may be harmful to the images' quality.

 
\subsection{The Change of Energy in Generation}
Image generation by an original robust classifier is introduced in the main paper, and the generating procedure of images can be formulated as: 
\begin{equation}
    x' = x - \frac{\eta}{2} \cdot\nabla_{x}(E_{\theta}(x,y)-E_{\theta}(x)) +\sqrt{\eta}\epsilon.
\label{generateeq}
\end{equation}

\begin{figure}[ht!]
 \vskip 0.2in
 \begin{center}  
 \centerline{\includegraphics[width=1.0\linewidth]{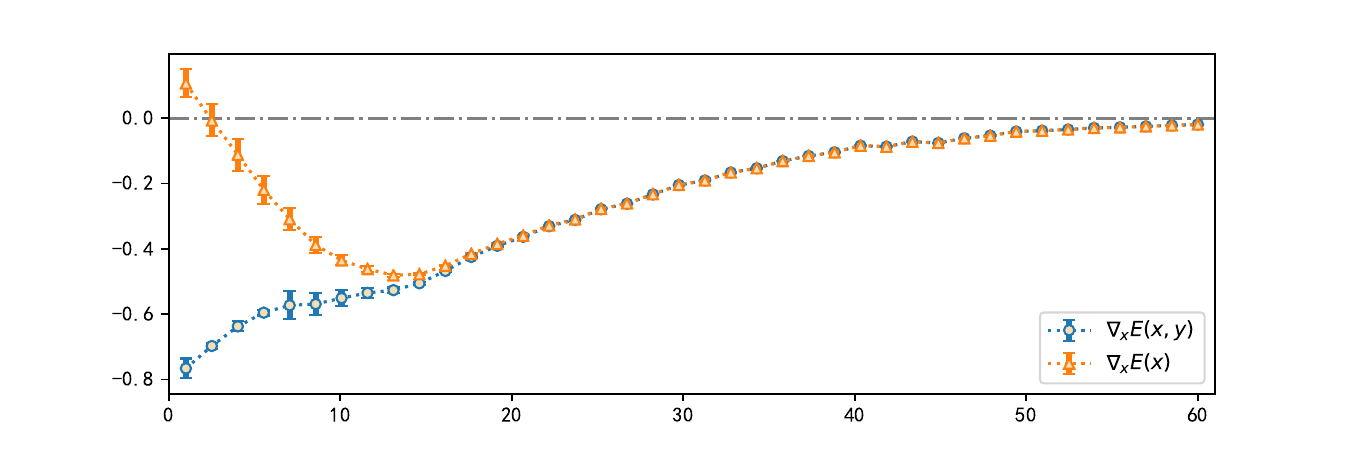}}
 \caption{$E_\theta(x,y)$ plays a major role when generating images by the original robust classifier (CIFAR-10). We illustrate the $\nabla_{x}E_{\theta}(x,y)$ and $\nabla_{x}E_{\theta}(x)$ in the generation process of a robust classifier (WRN-28-10) on CIFAR-10. The center points of the tags represent the mean value and the lengths represent the variance. The horizontal axis represents the number of iterations.}
 \label{fig:delta_energy_generation}
 \end{center}
 \vskip -0.2in
 \end{figure}

\begin{figure}[ht!]
 \vskip 0.2in
 \begin{center}  
 \centerline{\includegraphics[width=1.0\linewidth]{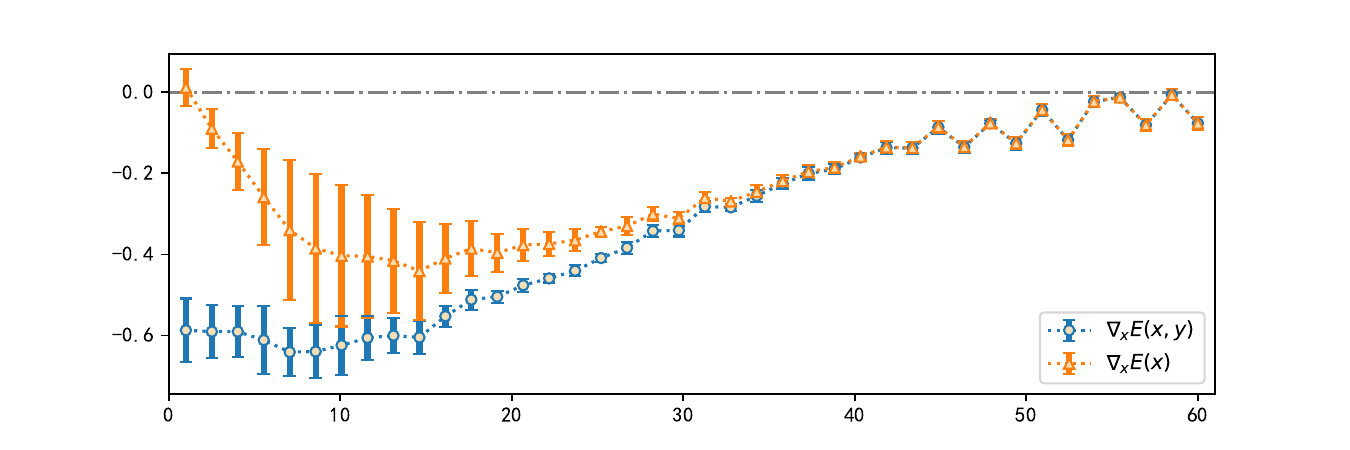}}
 \caption{$E_\theta(x,y)$ plays a major role when generating images by the original robust classifier (CIFAR-100). We illustrate the $\nabla_{x}E_{\theta}(x,y)$ and $\nabla_{x}E_{\theta}(x)$ in the generation process of a robust classifier (WRN-28-10) on CIFAR-100. The center points of the tags represent the mean value and the lengths represent the variance. The horizontal axis represents the number of iterations.}
 \label{fig:delta_energy_generation_cifat100}
 \end{center}
 \vskip -0.2in
 \end{figure}

As analyzed in Sec. 3.2, $E_{\theta}(x,y)$ is the key term for conditional generation by using robust classifier. 
We illustrate $\nabla_{x}E_{\theta}(x,y)$ and $\nabla_{x}E_{\theta}(x)$ in generation process in Fig \ref{fig:delta_energy_generation} (CIFAR-10) and Fig. \ref{fig:delta_energy_generation_cifat100} (CIFAR-100). In the beginning, $\nabla_{x}E_{\theta}(x,y)$ dominates for that the absolute value of $\nabla_{x}E_{\theta}(x)$ is less than that of $\nabla_{x}E_{\theta}(x,y)$. As the number of iterations increases, the influence of $\nabla_{x}E_{\theta}(x)$ and $\nabla_{x}E_{\theta}(x,y)$ gradually become much more closer. Thus, $E_{\theta}(x,y)$ plays a key role for conditional generation by using robust classifier. 

\section{The Gradient on $\theta$ of $\log p_\theta(x)$}
$p_\theta(x)$ in JEAT algorithm can be expressed as: 
 \begin{equation}
 p_{\theta}(x) = \frac{\exp(-E_{\theta}(x))}{Z_{\theta}},
 \end{equation}
 The gradient on $\theta$ of $\log p_\theta(x)$ is defined as
 
\begin{equation}
    \begin{array}{ll}
    \nabla_{\theta} \log(p_{\theta}(x)) & = -  \nabla_{\theta}E_\theta(x) - \frac{1}{Z_\theta} \nabla_{\theta} Z_\theta \\
    
    &= - \nabla_{\theta} E_\theta(x) - \frac{1}{Z_\theta} \nabla_{\theta}  \int_{x}\exp(-E_{\theta}(x))dx \\
     &= - \nabla_{\theta} E_{\theta}(x) + \int_{x}\frac{\exp(-E_{\theta}(x)}{Z_\theta} \nabla_{\theta}  
     E_{\theta}(x) dx \\
     
     & = - \nabla_{\theta} E_{\theta}(x) + \mathbb{E}_{p_{\theta}(x)}(\nabla_{\theta} E_{\theta}(x))
    \end{array}
    \label{gpx2}
\end{equation}
 We use Stochastic Gradient Langevin Dynamics (SGLD) to approximate $\mathbb{E}_{p_{\theta}(x)}$.

\section{Proof for $Z_\theta=\Tilde{Z}_\theta$}
As defined in the main paper, the energy functions for classifier are:
\begin{equation}
\left\{
\begin{array}{l}
E_{\theta}(x,y) = -\log(\exp({f(x;\theta)[y]})), \\
E_{\theta}(x) = -\log(\sum_{k=1}^{n} \exp({f(x;\theta)[k]})). 
\end{array}
\right.
\label{energy_x2}
\end{equation}

Also $p_{\theta}(x,y) =  \frac{\exp(-E_{\theta}(x,y))}{\Tilde{Z_{\theta}}}$, $\Tilde{Z_{\theta}}$ is the normalizing constant which is defined as: 
\begin{equation}
\begin{array}{ll}
\Tilde{Z_{\theta}} & = \int_{x}\sum\limits_{y}\exp(-E_{\theta}(x,y))dx\\
           & = \int_{x}\sum\limits_{y}\exp({f(x;\theta)[y]})dx 
\end{array}        
\label{zbar}
\end{equation}

As $p_{\theta}(x) =  \frac{\exp(-E_{\theta}(x))}{Z_{\theta}}$, $Z_{\theta}$ is the normalizing constant which is defined as: 
\begin{equation}
\begin{array}{ll}
Z_{\theta} & = \int_{x}\exp(-E_{\theta}(x))dx\\
           & = \int_{x}\exp(\log(\sum\limits_{y} \exp({f(x;\theta)[y]})))dx\\
           & = \int_{x}(\sum\limits_{y} \exp({f(x;\theta)[y]}))dx           
\end{array}  
\label{ztheta}
\end{equation}

By Eq. (\ref{zbar}) and Eq. (\ref{ztheta}), it can be seen that $Z_\theta=\Tilde{Z}_\theta$.

 \begin{algorithm}[t]
    \vskip 0.2in    
    \caption{Training and Generating of PreJEAT: Given network $f$,  $E(x,y)$ = $-\log(\exp({f(x)[y]}))$ represent energy of $(x,y)$,  $E(x)$ = $-\log(\sum_{y} \exp({f(x)[y]}))$ represent energy of $x$, $\ell_{\infty}$ adversarial perturbation radius $\epsilon$, SGLD step-size $\alpha$, SGLD steps $K$, epochs $T$, dataset of size $M$, learning rate $\eta$.}
    \begin{algorithmic}
    \State {\textbf{Training}}: 
    \For {$i = 1,2...,T$}
        \For {$j = 1,2...,M$} 
            \State $\blacktriangleright$ Generating energy-based adversarial samples:
            \State $\delta=\mathcal{U}(-\epsilon,\epsilon)$
            \State $\delta=\delta+\epsilon \cdot sign(\nabla_{x}(E_{\theta}(x,y)))$
            \State $\delta=max(min(\delta,\epsilon),-\epsilon)$
            \State $x_{adv}=x_j+\delta$

            \State $\blacktriangleright$ Updating model parameters:
            \State $\nabla_{\theta}\mathcal{L}_{p_\theta(y|x_{adv})}$=$\nabla_{\theta} (E_{\theta}(x_{adv},y) - E_{\theta}(x_{adv}))$
            \State $\theta=\theta-\eta \cdot \nabla_{\theta}\mathcal{L}_{p(y|x_{adv})}$
        \EndFor
	\EndFor

   \State
   \State
    {\textbf{Generating}}:
        \State $x_{0}$ $\sim$ \text{random sample}
        \For {$t = 0,1,2...,K-1$}
            \State
            $x_{t+1} = x_{t} - \frac{\alpha}{2}\cdot\nabla_{x_{t}}E_{\theta}(x_{t},y) + \sqrt{\alpha} \cdot \mathcal{N}(0,I)$
		\EndFor
		\State {\textbf{Output}:} $x_{gen}=x_{K}$
    \end{algorithmic}
    \label{algorithm:PreJEAT-algorithm}
\end{algorithm}

\section{Algorithm for PreJEAT}
In the main paper, we show the training algorithm called Joint Energy Adversarial Training (JEAT), in which we replace cross-entropy loss $-\log p_\theta(y|x)$ with $-\log p_\theta(x,y) =-\log p_\theta(y|x)-\log p_\theta(x)$ and use adversarial examples found by 
\begin{equation}
     x_{adv} = x + \epsilon \cdot sign(\nabla_{x}(E_{\theta}(x,y))).
\label{xadv32}
\end{equation}
We also propose Preliminary Joint Energy Adversarial Training (PreJEAT) which just trains model with adversarial examples found by Eq. (\ref{xadv3}) and use cross-entropy loss. We present the algorithm here in Algorithm \ref{algorithm:PreJEAT-algorithm}.

\section{Training Details}
The hyper-parameters for our experiments are shown in Tab.~\ref{hyperparameters}. We run JEM \cite{grathwohl2020classifier}, Free adversarial training (Free m=8) \cite{shafahi2019adversarial}, Fast adversarial training (Fast) \cite{wong2020fast} and TRADES (1/$\lambda$=6)\cite{tradesoff} with
their open-source code. MultiStep scheduler decay by 10 every time.

\section{Interesting Benefits of JEAT}
\subsection{Denoise}
Due to the influence of factors such as the environment and transmission channels, the image is inevitably contaminated by noise in the process of acquisition, compression and transmission. In the presence of noise, subsequent image processing tasks (such as video processing, image analysis, and tracking) may be negatively affected. Therefore, image denoising plays an important role in modern image processing systems.
In this section, we found that the classifier trained by JEAT has the effect of denoising to some extent.

\begin{figure}[ht!]
 \vskip 0.2in
 \begin{center}  
 \centerline{\includegraphics[width=1.0\linewidth]{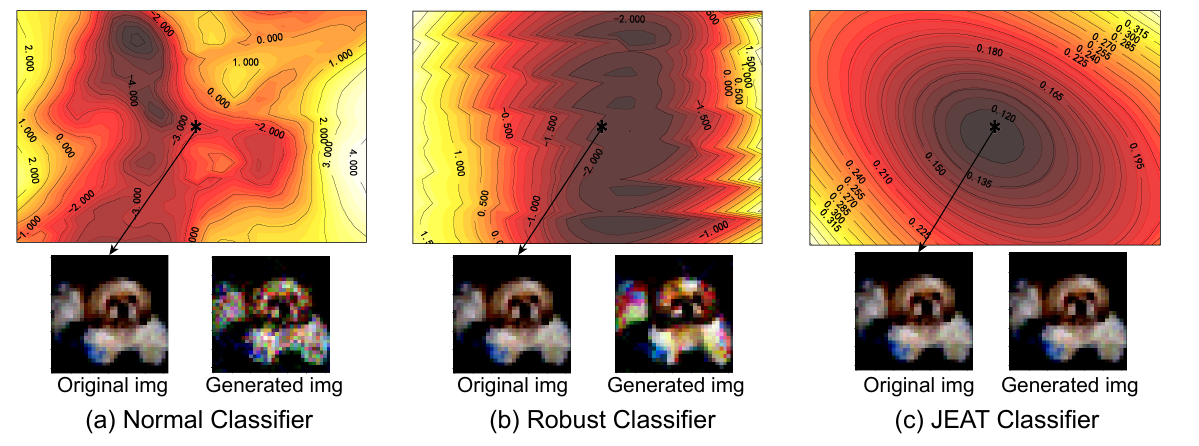}}
 \caption{We plot the energy contour around a given image of normal classifier (a), robust classifier (b), and JEAT classifier (c). The center point represents the energy of the given image. Darker colors in the diagram correspond to lower energy. Starting from the original image, we generate images by different classifiers using Langevin Dynamics.}
 \label{fig:energy_three_models}
 \end{center}
 \vskip -0.2in
 \end{figure}

\begin{figure}[ht!]
 \vskip 0.2in
 \begin{center}  
 \centerline{\includegraphics[width=1.0\linewidth]{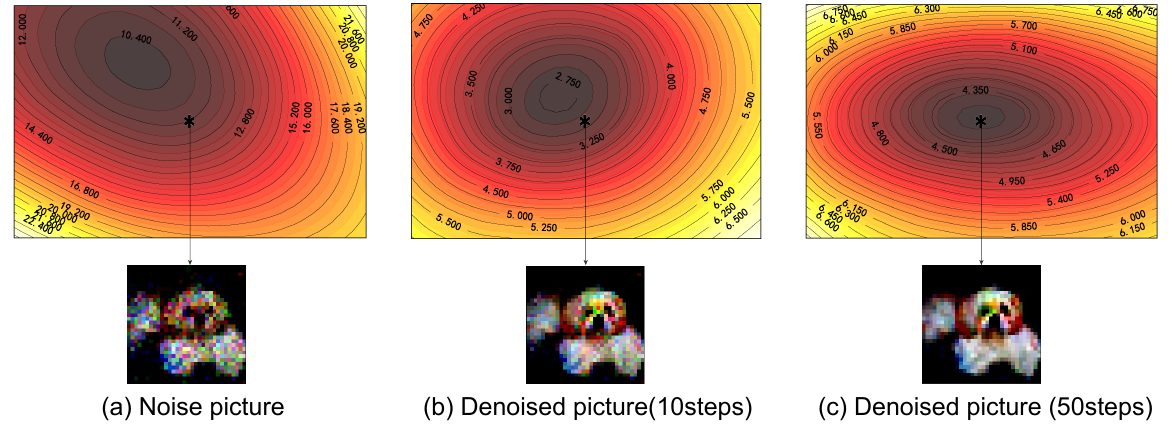}}
 \caption{We show the change of energy contour in the process of denoising. The initial noisy image has higher energy, and then the denoised image is gradually obtained along the direction of energy descent.}
 \label{fig:during_denoise}
 \end{center}
 \vskip -0.2in
 \end{figure}

\begin{figure}[ht!]
 \vskip 0.2in
 \begin{center}  
 \centerline{\includegraphics[width=1.0\linewidth]{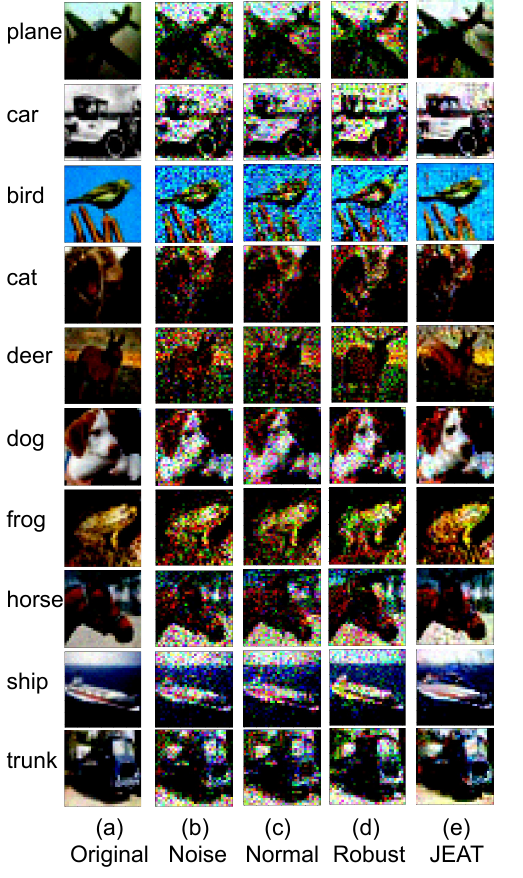}}
 \caption{We show that compared to the normal classifier and original robust classifier, JEAT classifier can indeed achieve the effect of denoising. (a) Original images. (b) Noise images obtained by superimposing Gaussian noise with a mean value of 0 and a variance of 0.3 on the original image. (c) The image obtained after denoising by the normal classifier. (d) The image obtained after denoising by the robust classifier. (e) The image obtained after denoising by the JEAT classifier.}
 \label{fig:denoise}
 \end{center}
 \vskip -0.2in
 \end{figure}
We further show the energy contours of the normal classifier, robust classifier and JEAT classifier around a given image from test dataset (CIFAR-10) in Fig. \ref{fig:energy_three_models}. The points with the lowest energy of both the normal classifier and the robust classifier deviate from the point where the real image is located. The generation process will result in entering the low energy positions along the direction of energy descent from the real image. Therefore, even if it starts from the real image, the image generated by the normal classifier will be inferior. Because the low-energy region of the original robust classifier deviates slightly from the real data, the quality of generated images will be slightly better than that of the normal classifier. 
Nevertheless, in the energy of JEAT classifier, the real image has the minimum energy, and the energy function is flat and smooth around the real data. Starting from the real image, the image generated by the JEAT classifier is almost identical to the real image. 

This inspires us that if the energy of the noise image is high, it is possible to generate an clean image along the direction of energy descent, thereby denoising. We show the different energy contours in the process of denoising by JEAT classifier in Fig. \ref{fig:during_denoise}. We get noisy images by injecting Gaussian noise with a mean value of zero and a variance of 0.3 into the original image. Then we generate low-energy images from these noisy images using Langevin Dynamics. As shown in Fig. \ref{fig:denoise}, JEAT classifier has good denoising effect, while normal classifier and original robust classifier have poor denoising effect.

\subsection{Calibration}

\begin{figure}[ht!]
 \vskip 0.2in
 \centering
 \centerline{\includegraphics[width=1.0\linewidth]{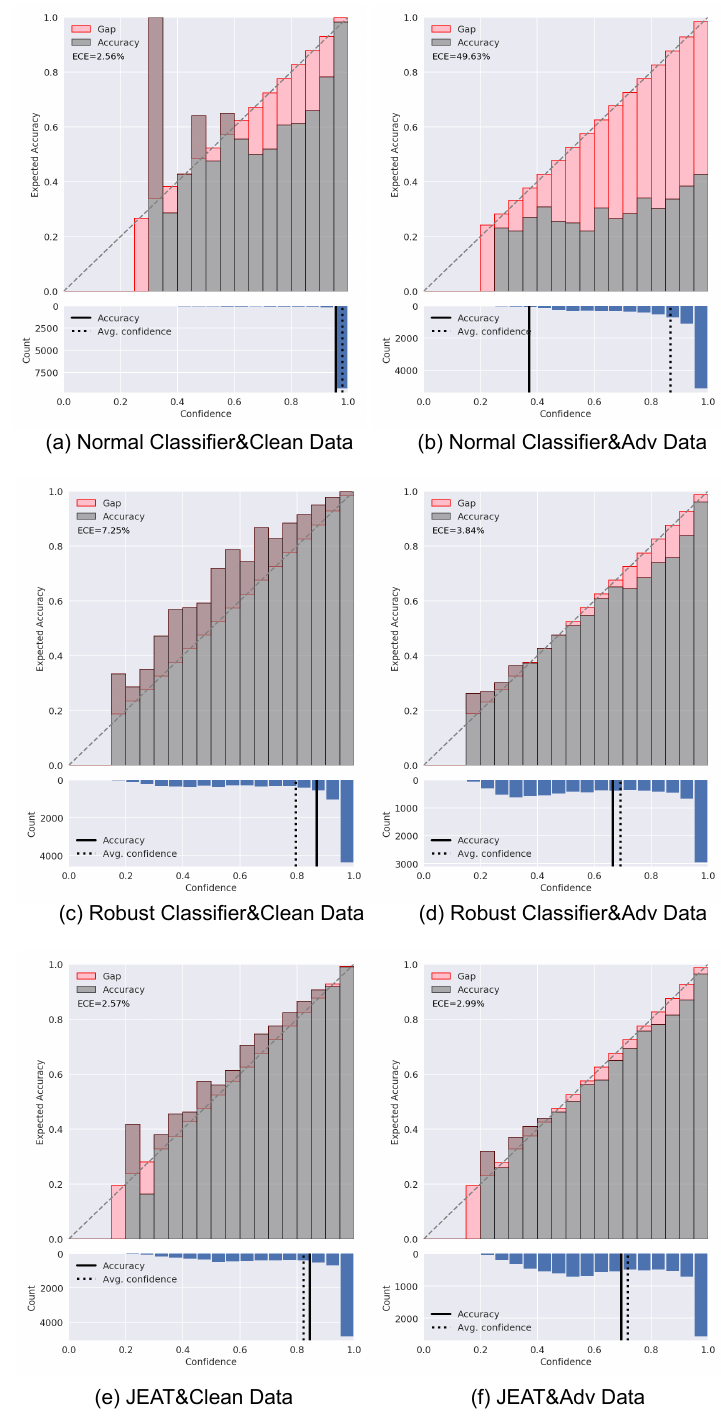}}
 \caption{We draw the reliability diagram of different classifiers on original test data of CIFAR-10 and adversarial data. We use FGSM attack ($\epsilon=8/255$)~\cite{goodfellow2014explaining} to get adversarial data. (a) The reliability diagram of normal classifier on original clean data. (b) The reliability diagram of normal classifier on adversarial data. (c) The reliability diagram of robust classifier on original clean data. (d) The reliability diagram of robust classifier on adversarial data. (e) The reliability diagram of JEAT classifier on original clean data. (f) The reliability diagram of JEAT classifier on adversarial data. }
 \label{fig:calibration}
 \vskip -0.2in
 \end{figure}
 
The outputs of a classifier are often interpreted as the predictive confidence that this class was identified. However Guo et al.~\cite{guo2017calibration} claim that deep neural networks are often not calibrated which means that the confidence always does not align with misclassification rate. Expected Calibration Error (ECE) is a metric to measure the calibration of a classifier. For a perfectly calibrated classifier, ECE value will be zero. 

We using the reliability diagram to find out how well the classifier is calibrated in Fig. ~\ref{fig:calibration}. The model's predictions are divided into bins based on the confidence value of the target class, here, we choose 20 bins. The confidence histogram at the bottom shows how many test examples are in each bin. Two vertical lines represent accuracy and average confidence, and the closer these two lines are, the better the model calibration is. The reliability histogram at the top shows the average confidence of each bar and the accuracy of the examples in the bar. For each bin we plot the difference between the accuracy and the confidence using the red bars in the diagram.

It can be clearly seen from the histogram that the confidence of most predictions of these classifiers is greater than 0.8. The normal classifier is always over-confidence and gives more false positives. This phenomenon becomes even worse when facing adversarial data and the ECE value is 49.63$\%$. Robust classifier will not give much over-confident prediction when encountering adversarial perturbation. However, when faced with clean data, the robust classifier is under-confidence and gives more false negatives. The good news is that the JEAT classifier has a good calibration when classifying clean data or adversarial data. When a model is deployed in a real-world scenarios, good calibration is an important feature. And the confidence of a model with good calibration can be used to judge whether to output the result or recognize it again. The JEAT classifier which is better calibrated than both normal classifier and original robust classifier can be more useful in real-world scenarios.

\end{document}